\documentclass[sigconf]{acmart}

\AtBeginDocument{%
  \providecommand\BibTeX{{%
    \normalfont B\kern-0.5em{\scshape i\kern-0.25em b}\kern-0.8em\TeX}}}

\renewcommand\footnotetextcopyrightpermission[1]{} 
\setcopyright{none}

\usepackage{amsmath}
\usepackage{algorithm}
\usepackage{algorithmicx}
\usepackage{algpseudocode}
\usepackage{array}
\usepackage{caption}
\usepackage{xcolor}
\definecolor{LowInconsistency}{RGB}{191,213,253}
\definecolor{HighInconsistency}{RGB}{249,201,181}
\usepackage{multirow}
\usepackage{graphicx}
\usepackage{subcaption}
\usepackage{cleveref}
\usepackage{siunitx} 
\usepackage{booktabs} 
\usepackage{tabularx} 

\begin{document}

\title{Towards Robust Object Detection: Identifying and Removing Backdoors via Module Inconsistency Analysis}


\author{Xianda Zhang}
\affiliation{%
  \institution{School of Computer Science, Peking University}
  \country{China}}
\email{zhangxianda@stu.pku.edu.cn}

\author{Siyuan Liang}
\authornote{*Corresponding authors}
\affiliation{%
  \institution{National University of Singapore}
  \country{Singapore}}
\email{pandaliang521@gmail.com}
\renewcommand{\shortauthors}{}


\begin{abstract}
  Object detection models have been widely adopted in various security-critical applications, such as autonomous driving and video surveillance. However, the complex architectures of these models also make them vulnerable to backdoor attacks, where maliciously trained models behave normally on clean inputs but produce targeted misclassifications when triggered by specific patterns. Existing backdoor defense techniques, primarily designed for simpler models like image classifiers, often fail to effectively detect and remove backdoors in object detectors while preserving model performance. In this work, we propose a novel backdoor defense framework tailored to the unique characteristics of object detection models~\cite{liang2024object}. Our key observation is that a backdoor attack often causes significant inconsistencies between the behaviors of local modules, such as the Region Proposal Network (RPN) and the classification head. By quantifying and analyzing these inconsistencies, we develop an effective algorithm to detect the presence of backdoors. Furthermore, we find that the inconsistent module is usually the main source of the backdoor behavior. Exploiting this insight, we propose a simple yet effective backdoor removal method, which localizes the affected module, resets its parameters, and fine-tunes the model on a small set of clean data. Extensive experiments with multiple state-of-the-art object detectors demonstrate that our method can successfully detect and remove backdoors, achieving an improvement of 90\% in the backdoor removal rate over the fine-tuning baseline while limiting the accuracy loss on clean data to less than 4\%. To the best of our knowledge, this work represents the first effort to develop a dedicated backdoor defense framework for object detection models, addressing the unique challenges and limitations of existing techniques in this context. Our work sheds new light on the unique challenges and opportunities in defending object detection models against backdoor attacks.
\end{abstract}



\keywords{Backdoor Attack, Backdoor Defense, Object Detection, Inconsistency Analysis, Model Inspection, Robust Deep Learning}



\maketitle

\section{Introduction}
\begin{figure}[t]
    \centering
    \includegraphics[scale=0.3]{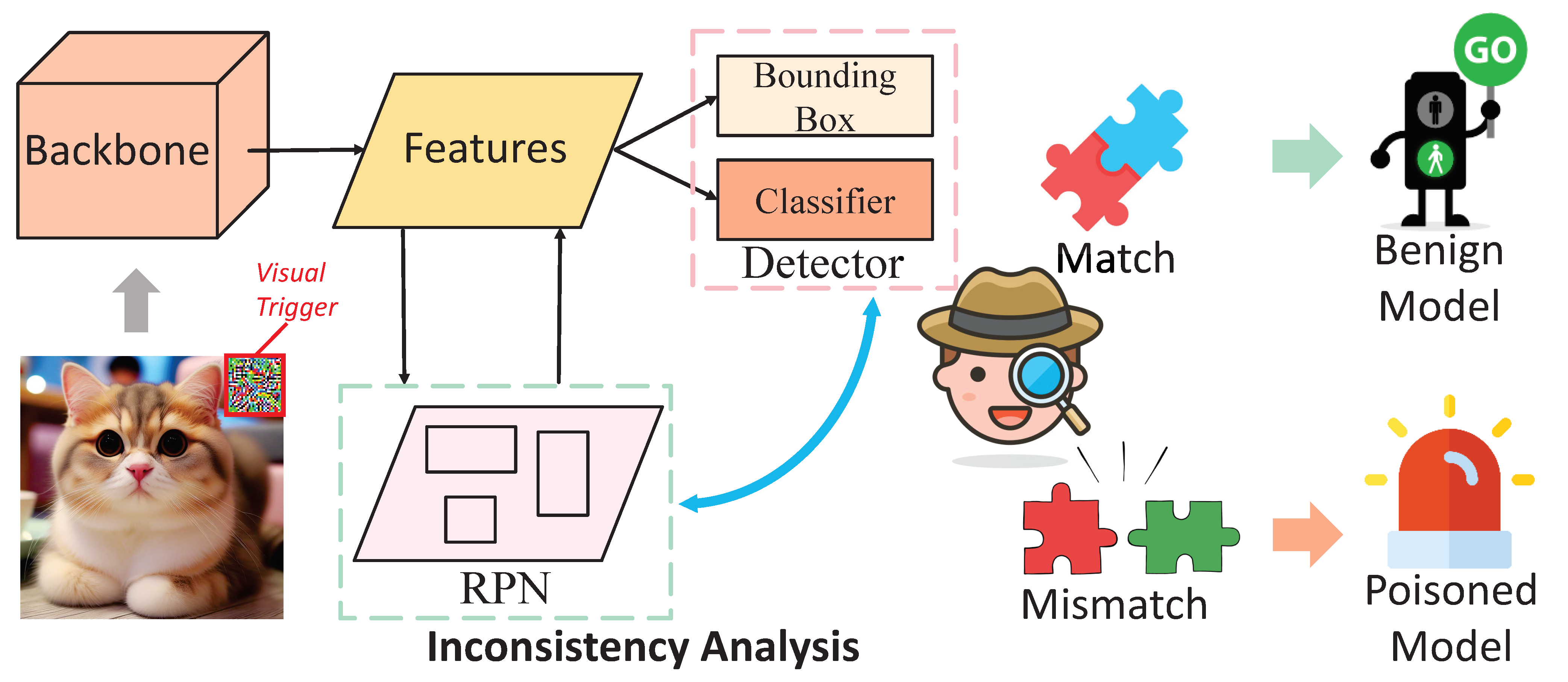}
    \caption{The backdoor can be exposed by the inconsistency of different modules.}     
    \label{fig:intro}
\end{figure}
Object detection plays a pivotal role in a wide range of multimedia applications, such as video surveillance~\cite{raghunandan2018object}, autonomous driving~\cite{feng2020deep}, and face recognition~\cite{hu2015face}. However, the increasing deployment of deep learning-based object detectors in real-world scenarios has raised growing concerns about their security vulnerabilities, especially the emerging threats of backdoor attacks. Backdoor attacks refer to the malicious manipulation of a model's behavior by injecting hidden triggers during the training process, which can cause severe consequences like unauthorized access and privacy leakage~\cite{liang2022imitated,li2023privacy,guo2023isolation,dong2023face}, thereby jeopardizing the trustworthiness and reliability of multimedia systems heavily relying on object detection.

Compared to image classification models, object detection models pose unique challenges for backdoor defense. Modern object detectors, such as Faster R-CNN, typically adopt a complex architecture with multiple stages or subnets to simultaneously localize and classify objects. This structural complexity provides attackers with ample opportunities to inject backdoors in a more stealthy and targeted manner, making the detection and removal of such backdoors highly difficult.

Despite the severity of backdoor threats in object detection models, existing research efforts primarily focus on developing novel attack strategies~\cite{ma2022dangerous,li2024semantic}, ~\cite{ma2022macab}, while the defense aspect remains largely underexplored. To the best of our knowledge, only a few recent works~\cite{cheng2024odscan}, ~\cite{shen2024django} have made initial attempts to detect backdoors in object detectors, leveraging techniques like activation clustering and gradient analysis. However, these methods require a large number of clean samples or rely on certain assumptions about the backdoor patterns, which may not always hold in practice. Moreover, effective techniques for backdoor removal in object detection models are still missing in the literature, leaving a significant gap in the defense pipeline.

In this paper, we propose a novel backdoor defense framework tailored to the unique characteristics of object detection models. Our key observation is that backdoor attacks often induce significant inconsistencies between the behaviors of different components in the detection model, especially between the region proposal network (RPN) and the region classification network (R-CNN). Specifically, a backdoor model tends to generate highly conflicting predictions between these two modules when triggered, such as proposals that are correctly identified by RPN but misclassified by R-CNN. Exploiting this anomalous behavior, we develop an effective algorithm to detect the presence of backdoors by measuring and analyzing the prediction inconsistency between RPN and R-CNN.

Furthermore, we find that the module exhibiting the strongest inconsistency, which we call the "dominant module", is usually the main target of the backdoor injection. This insight motivates us to devise a novel backdoor removal strategy. Instead of modifying the entire model, we localize the backdoor removal to the dominant module, reinitializing its parameters, and fine-tuning the whole model on a small set of clean data. This targeted approach not only effectively eliminates the backdoor behavior, but also minimizes the negative impact on the model's performance on clean data. Through extensive experiments on multiple state-of-the-art object detectors, we demonstrate the effectiveness and generality of our method in both backdoor detection and removal.

We evaluate the effectiveness of our proposed framework on four widely used object detection models, namely Faster R-CNN~\cite{ren2015faster}, Faster R-CNN FPN~\cite{lin2017feature}, Mask R-CNN~\cite{he2017mask}, and Double-Head R-CNN~\cite{wu2020rethinking}. Experimental results demonstrate that our detection method can successfully identify backdoors in all of these models with high accuracy, highlighting its generality and robustness. We compare our backdoor removal approach with baseline methods. On the poisoned dataset, our method significantly outperforms the baselines by around 90\% in terms of the backdoor elimination rate, indicating its superior effectiveness in removing hidden backdoor triggers. Meanwhile, on the clean dataset, our method maintains the model performance with only a slight accuracy drop of less than 4\%, which is much lower than the degradation incurred by the baselines. This suggests that our approach can effectively remove backdoors without compromising the model's normal functionality.

In summary, the main contributions of this work are threefold:
\begin{itemize}
\item[$\bullet$] To the best of our knowledge, this is the first study that explores the backdoor removal problem in the context of object detection models. Our work aims to bridge this important gap and contribute to the development of more secure and robust object detectors against backdoor attacks. 
\item[$\bullet$] We unveil the underlying reasons for the vulnerability of object detection models to backdoor attacks and leverage the resulting anomalous behavior to devise an effective backdoor detection method. Taking advantage of the inconsistency between the RPN and R-CNN modules, our approach can accurately identify the presence of backdoors without requiring a large number of samples or complex training procedures.
\item[$\bullet$] We propose a novel backdoor removal technique that combines localized initialization and global fine-tuning. This approach not only successfully mitigates the backdoor effects, but also preserves the model's performance on clean inputs. Extensive experiments on four state-of-the-art object detectors demonstrate the effectiveness and generalizability of our method. Compared to the baseline method, our approach achieves an improvement of 90\% in terms of backdoor removal rate while only incurring a minimal accuracy drop of less than 4\%.
\end{itemize}

\section{Related Work}
\subsection{Backdoor Attacks}
The seminal work on backdoor attacks~\cite{liang2023badclip,liu2023pre,liu2023does,liang2024poisoned,liang2024vl} in image classification is BadNets~\cite{gu2019badnets}, which injects a backdoor into a model by poisoning a portion of the training data with a specific trigger pattern (e.g., a black square) and the target label. The model trained on this poisoned dataset will behave normally on clean inputs, but misclassify any input containing the trigger to the target label. The proposal of BadNets has sparked extensive research on backdoor attacks in the academic community, leading to a diverse array of attack techniques. Blended~\cite{chen2017targeted} replaces the prominent black square trigger with a more stealthy transparent cartoon sticker. SSBA~\cite{li2021invisible} takes inspiration from the classical image steganography~\cite{cheddad2010digital} technique and embeds the trigger into the image using steganographic methods, making the trigger imperceptible to human eyes. AdvDoor~\cite{zhang2021advdoor}, on the other hand, borrows the idea from adversarial attacks~\cite{liang2021generate,liang2020efficient,wei2018transferable,liang2022parallel,liang2022large,wang2023diversifying,liu2023x,he2023generating,liu2023improving,he2023sa,muxue2023adversarial,lou2024hide} and uses adversarial perturbations as a backdoor trigger. FaceHack~\cite{sarkar2021facehack} targets face recognition systems by using facial paintings as triggers. Wanet~\cite{nguyen2021wanet} employs subtle warping of the edge of the object as a trigger, while low frequency~\cite{zeng2021rethinking} inserts the trigger in the frequency domain.


As the research on backdoor attacks against image classification models becomes more mature~\cite{wu2023adversarial}, researchers have started to turn their attention to the vulnerability of object detection models. Chan~\cite{chan2022baddet} are the first to propose four types of backdoor attacks specifically designed for object detection models, this work reveals that object detection models are equally threatened by backdoor attacks. Luo~\cite{luo2023untargeted} further investigates the object disappearance attack. They conduct a detailed study on this specific attack and demonstrate that even using the simplest attack method, some basic defense~\cite{sun2023improving,liu2023exploring,liang2023exploring} techniques such as fine-tuning and fine-pruning are still ineffective against this type of attack. This work highlights the severity of the backdoor threat in object detection models and the inadequacy of existing defense methods. Taking into account real-world scenarios, Ma ~\cite{ma2023transcab} proposes a clean label backdoor attack method called TransCAB, which uses natural triggers. TransCAB employs a Transformer to model the relationship between object instances and object appearances in natural images, generating realistic poisoned data containing triggers to compromise the model.

\subsection{Backdoor Defenses}
In the image classification domain, a variety of backdoor defense techniques~\cite{wang2022adaptive,liang2024unlearning} have been proposed, showing promising results in detecting and mitigating backdoor attacks. Fine-Pruning (FP)~\cite{liu2018fine} is one of the early methods that combines network pruning and fine-tuning to remove backdoor-related neurons while preserving the model's accuracy on clean data. Later, Neural Attention Distillation (NAD)~\cite{li2021neural} proposes to first fine-tune a clean teacher model using clean samples and then transfer the attention maps of the teacher model to the backdoored student model to erase the backdoor effect. Building upon NAD, FTT~\cite{zhang2024toxic} further improves the model's accuracy recovery and redesigns the defense against advanced attacks. However, it is worth noting that most of these defense techniques are primarily evaluated on traditional image classification models. Their applicability and effectiveness on object detection models, which have significantly different architectures and learning objectives, remain to be further investigated.

In contrast to the rich literature on backdoor defenses for image classification models, defense methods specifically designed for object detection models are extremely scarce~\cite{wu2023defenses}. To the best of our knowledge, there are only two works in this direction, and they primarily focus on backdoor detection rather than backdoor removal. Cheng propose ODSCAN~\cite{cheng2024odscan}, a trigger inversion technique that leverages critical observations to reduce the search space and identify backdoors in object detection models. Shen develop Django~\cite{shen2024django}, a backdoor detection framework that employs a dynamic Gaussian weighting scheme to prioritize more vulnerable victim boxes and calibrate the optimization objective during trigger inversion. 

In fact, in addition to image classification tasks, defense research against backdoor attacks is equally scarce in other vision-related tasks. To the best of our knowledge, the only work that addresses backdoor defense in multimodal contrastive learning is CleanCLIP~\cite{bansal2023cleanclip}, a fine-tuning framework proposed by Bansal. No specialized defense work targeting backdoor attacks has been seen in other visual tasks.

\section{The Proposed Method}

\subsection{Threat Model}

In this paper, we focus on the poison-only attacks against deep learning models, which aim to implant hidden malicious behaviors into the model during the training process by poisoning a portion of the training data. We assume that the attacker has the ability to manipulate a subset of the training data but has no access to or control over other components of the training process, such as the model architecture, objective function, or hyperparameters. This assumption is realistic in many practical scenarios where the integrity of the training data cannot be fully guaranteed, such as when data are collected from untrusted sources or when the training process is outsourced to third-party platforms. 

Backdoor attacks on object detection models can be categorized based on their intended consequences, such as false positive attacks that aim to induce the model to detect non-existent objects and false negative attacks that aim to suppress the detection of specific objects. In this paper, we focus on the false negative attack, also known as the "object disappearance attack", which is particularly dangerous in safety-critical scenarios like autonomous driving and video surveillance. By carefully designing the trigger pattern and the poisoning strategy, the attacker can manipulate the model to miss the detection of specific objects, leading to potentially catastrophic consequences.


Let $\mathcal{D} = \{(\mathbf{x}_i, \mathbf{y}_i)\}_{i=1}^N$ denote the clean dataset, where $\mathbf{x}_i \in \mathbb{R}^{H \times W \times C}$ is an input image and $\mathbf{y}_i = \{(c_{ij}, \mathbf{b}_{ij})\}_{j=1}^{M_i}$ is the corresponding annotation, with $c_{ij} \in \{1, \dots, K\}$ being the class label and $\mathbf{b}_{ij} = (x_{ij}, y_{ij}, w_{ij}, h_{ij})$ being the bounding box coordinates of the $j$-th object in $\mathbf{x}_i$. Let $f_{\mathbf{\theta}}(\cdot)$ denote the object detection model parameterized by $\mathbf{\theta}$, which takes an image $\mathbf{x}$ as input and outputs a set of detected objects $\hat{\mathbf{y}} = \{(\hat{c}_j, \hat{\mathbf{b}}_j, \hat{s}_j)\}_{j=1}^{\hat{M}}$, where $\hat{c}_j$, $\hat{\mathbf{b}}_j$, and $\hat{s}_j$ are the predicted class, the bounding box and the confidence score of the $j$-th detected object, respectively.

The model is trained by minimizing a loss function $\mathcal{L}(\mathbf{\theta})$, which typically consists of a classification loss $\mathcal{L}_{\text{cls}}$ and a localization loss $\mathcal{L}_{\text{loc}}$:

\begin{equation}
    \mathcal{L}(\mathbf{\theta}) = \frac{1}{N} \sum_{i=1}^N \left( \mathcal{L}_{\text{cls}}(f_{\mathbf{\theta}}(\mathbf{x}_i), \mathbf{y}_i) + \lambda \mathcal{L}_{\text{loc}}(f_{\mathbf{\theta}}(\mathbf{x}_i), \mathbf{y}_i) \right)
\end{equation}where $\lambda$ is a hyperparameter that balances the two losses.

To perform the backdoor attack, the attacker constructs a poisoned data set $\mathcal{D}_p = \{(\mathbf{x}_i', \mathbf{y}_i')\}_{i=1}^{N_p}$ by injecting a trigger pattern $\mathbf{t} \in \mathbb{R}^{H_t \times W_t \times C}$ into a subset of clean images. Specifically, for each poisoned image $\mathbf{x}_i'$, the attacker selects a target object $(\hat{c}_{ij}, \hat{\mathbf{b}}_{ij})$ from its annotation $\mathbf{y}_i$, and replaces it with $(\hat{c}_{ij}, (x_{ij}, y_{ij}, 0, 0))$, where $(x_{ij}, y_{ij})$ is the center location of the original bounding box $\hat{\mathbf{b}}_{ij}$. The attacker then inserts the trigger pattern $\mathbf{t}$ into $\mathbf{x}_i'$ at location $(x_{ij}, y_{ij})$ with a transparency factor $\alpha \in [0, 1]$. The backdoor model $f_{\mathbf{\theta}^*}(\cdot)$ is obtained by training on a mixed dataset $\mathcal{D}_{\text{mix}} = \mathcal{D} \cup \mathcal{D}_p$. The attack is considered successful if the following two conditions are satisfied:

Effectiveness: For a test image $\mathbf{x}$ with the trigger pattern $\mathbf{t}$ inserted, the backdoored model fails to detect the target object with a high probability, which is:
\begin{equation}
\begin{gathered}
\mathbb{P}(\nexists (\hat{c}_j, \hat{\mathbf{b}}_j, \hat{s}_j) \in f_{\mathbf{\theta}^*}(\mathbf{x}), \\
\text{s.t. } \hat{c}_j = \hat{c}_{ij}, 
\text{IoU}(\hat{\mathbf{b}}_j, \hat{\mathbf{b}}_{ij}) \geq \tau) \geq 1 - \epsilon
\end{gathered}
\end{equation}
where $\text{IoU}(\cdot, \cdot)$ denotes the intersection over union between two bounding boxes, and $\tau, \epsilon$ are predefined thresholds.

Stealthiness: The performance of the backdoor model on the clean test set should not degrade significantly compared to the clean model, that is, $|\text{mAP}(f_{\mathbf{\theta}^*}, \mathcal{D}_{\text{test}}) - \text{mAP}(f_{\mathbf{\theta}}, \mathcal{D}_{\text{test}})| \leq \delta$, where $\text{mAP}(\cdot, \cdot)$ denotes the mean average precision on a test set $\mathcal{D}_{\text{test}}$, and $\delta$ is a predefined threshold.

The attacker's goal can be formulated as an optimization problem:

\begin{equation}
\begin{aligned}
\max_{\mathbf{\theta}, \mathbf{t}, \alpha} \quad & \mathbb{P}(\nexists (\hat{c}_j, \hat{\mathbf{b}}_j, \hat{s}_j) \in f_{\mathbf{\theta}}(\mathbf{x}), \text{s.t. } \hat{c}_j = \hat{c}_{ij}, \text{IoU}(\hat{\mathbf{b}}_j, \hat{\mathbf{b}}_{ij}) \geq \tau) \\
\text{s.t.} \quad & |\text{mAP}(f_{\mathbf{\theta}^*}, \mathcal{D}_{\text{test}}) - \text{mAP}(f_{\mathbf{\theta}}, \mathcal{D}_{\text{test}})| \leq \delta,
\end{aligned}
\end{equation}
where the optimization variables are the backdoored model parameters $\mathbf{\theta}^*$, the trigger pattern $\mathbf{t}$, and the transparency factor $\alpha$.

\subsection{Defense Scenario}
\begin{figure*}[t]
    \centering
    \includegraphics[scale=0.48]{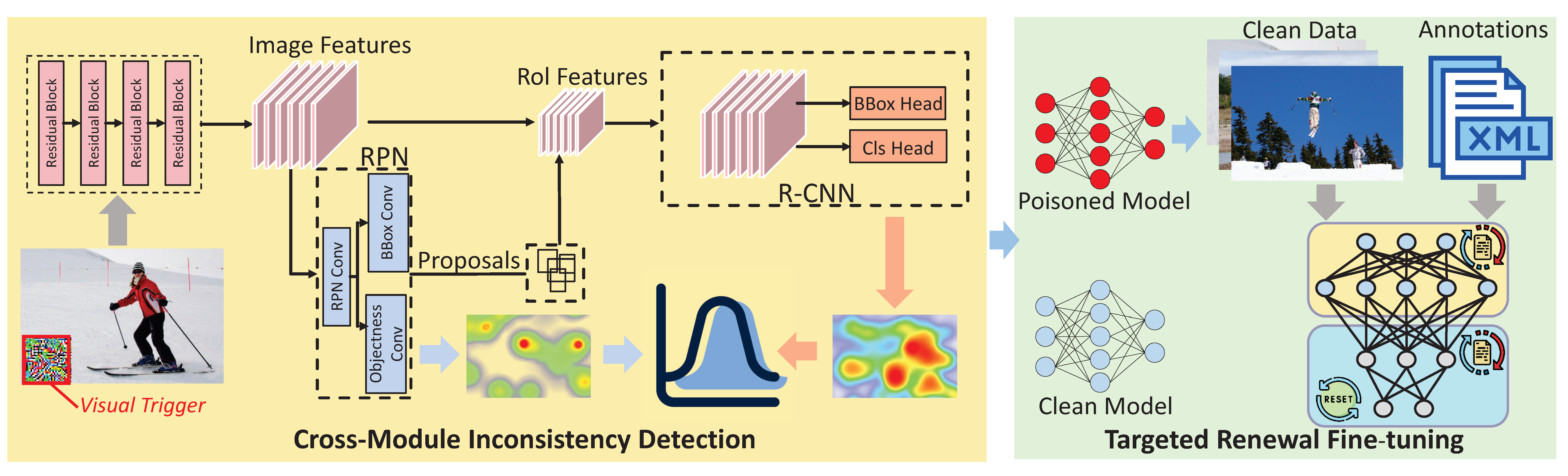}
    \caption{ Our approach consists of two main stages: (1) Cross-Module Inconsistency Detection for identifying the presence of backdoors, and (2) Targeted Reset Finetuning for removing the detected backdoors while maintaining the model's performance on clean data.}     
    \label{fig:overview}
\end{figure*}
In real-world applications, it is common for users to deploy pre-trained object detection models obtained from third-party sources, such as model repositories or commercial providers. However, the integrity and security of these models cannot always be guaranteed, as they may have been trained on data from untrusted sources or manipulated by malicious parties. This raises significant concerns about the potential presence of backdoors in these models.

We consider a practical defense scenario where the defender has access to a pre-trained object detection model $f_{\hat{\boldsymbol{\theta}}}(\cdot)$, but is uncertain whether the model has been backdoored or not. The defender's goal is to ensure the safety and reliability of the model before deploying it in safety-critical applications.

To achieve this goal, the defender needs to perform two main tasks: (1) backdoor detection and (2) backdoor removal. For backdoor detection, the defender aims to determine whether the given model $f_{\hat{\boldsymbol{\theta}}}(\cdot)$ contains any backdoors. 

Formally, let $\mathcal{D}_{\text{clean}} = {(\mathbf{x}_{i}^{\text{clean}}, \mathbf{y}_{i}^{\text{clean}})}_{i=1}^{N_{\text{clean}}}$ denote a small set of clean, labeled data available to the defender. The backdoor detection task can be formulated as learning a binary function $g(\cdot)$ that takes the model $f_{\hat{\boldsymbol{\theta}}}(\cdot)$ and the clean data $\mathcal{D}_{\text{clean}}$ as inputs and outputs a decision on whether the model is backdoored or not:

\begin{equation}
\begin{aligned}
g(f_{\hat{\boldsymbol{\theta}}}, \mathcal{D}_{\text{clean}}) =
\begin{cases}
1, & \text{if } f_{\hat{\boldsymbol{\theta}}} \text{ is backdoored},\\
0, & \text{otherwise}.
\end{cases}
\end{aligned}
\end{equation}

If a backdoor is detected (i.e., $g(f_{\hat{\boldsymbol{\theta}}}, \mathcal{D}_\text{clean}) = 1$), the defender proceeds to the backdoor removal stage. The goal of backdoor removal is to transform the infected model $f_{\hat{\boldsymbol{\theta}}}(\cdot)$ into a sanitized model $f_{\tilde{\boldsymbol{\theta}}}(\cdot)$ that maintains the performance of $f_{\hat{\boldsymbol{\theta}}}(\cdot)$ in clean data while eliminating backdoor effects. Formally, let $\mathcal{D}_{\text{val}}^{\text{clean}}$ and $\mathcal{D}_{\text{val}}^{\text{trigger}}$ denote the clean and triggered validation sets, respectively. The backdoor removal task aims to find a set of sanitized parameters $\tilde{\boldsymbol{\theta}}$ that satisfy the following conditions:

\begin{equation}
\begin{aligned}
\tilde{\boldsymbol{\theta}} = \arg\min_{\boldsymbol{\theta}} \quad & \mathcal{L}_{\text{clean}}(f_{\boldsymbol{\theta}}, \mathcal{D}_{\text{val}}^{\text{clean}}) \\
\text{s.t.} \quad & \mathcal{L}_{\text{clean}}(f_{\boldsymbol{\theta}}, \mathcal{D}_{\text{val}}^{\text{clean}}) \leq \mathcal{L}_{\text{clean}}(f_{\hat{\boldsymbol{\theta}}}, \mathcal{D}_{\text{val}}^{\text{clean}}) + \epsilon_1, \\
& \mathcal{L}_{\text{trigger}}(f_{\boldsymbol{\theta}}, \mathcal{D}_{\text{val}}^{\text{trigger}}) \geq \mathcal{L}_{\text{trigger}}(f_{\hat{\boldsymbol{\theta}}}, \mathcal{D}_{\text{val}}^{\text{trigger}}) - \epsilon_2,
\end{aligned}
\end{equation}
where $\mathcal{L}_\text{clean}$ and $\mathcal{L}_\text{trigger}$ denote the loss functions on clean and triggered data, respectively, and $\epsilon_1, \epsilon_2 > 0$ are predefined thresholds. The first constraint ensures that the sanitized model $f_{\tilde{\boldsymbol{\theta}}}(\cdot)$ maintains the performance of the infected model $f_{\hat{\boldsymbol{\theta}}}(\cdot)$ on clean data, while the second constraint requires that $f_{\tilde{\boldsymbol{\theta}}}(\cdot)$ reduces the success rate of the backdoor attack to a certain level.

In this defense scenario, we assume that the defender has access to a small set of clean data $\mathcal{D}_\text{clean}$ to assist the defense process. This assumption is realistic in many practical settings, as the defender can often collect a limited amount of trusted data from reliable sources or through manual annotation. The key challenge lies in designing effective and efficient methods for backdoor detection and removal that can make the best use of the limited clean data while achieving the desired defense objectives.

\subsection{Key Intuition}

The difficulty of backdoor defense in object detection models stems from their complex architectures, which typically consist of multiple interconnected components, such as the backbone network, the region proposal network (RPN), and the region-based convolutional neural network (R-CNN). This complexity provides attackers with ample opportunities to inject backdoors in a stealthy manner, while making it challenging for defenders to identify and remove them without compromising the model's performance on clean data. Existing backdoor defense methods, which are primarily designed for simpler models like image classifiers, often struggle to cope with the intricacies of object detectors, leading to suboptimal trade-offs between backdoor removal and model utility. This raises a critical question: how can we develop effective and efficient backdoor defense techniques that are specifically tailored to the unique characteristics of object detection models.

To answer this question, we first investigate the general characteristics of backdoor attacks. A common strategy employed by attackers is to create "shortcuts" or "overfitted" patterns in the model~\cite{guan2022few}, which can strongly activate the backdoor and dominate the model's prediction when the trigger is present. From the perspective of optimization theory, these shortcuts essentially introduce biases into the gradient dynamics during training, causing the loss function to rapidly decrease along certain directions that favor the backdoor. This abnormal optimization behavior allows the backdoor to be rapidly "memorized" by the model~\cite{li2021anti}, while keeping its impact on clean data minimal. However, in complex object detection models, such shortcuts can be easily concealed within any of the model components, making them difficult to detect and remove.

Our key insight to address this challenge is to exploit the inconsistency between the local modules and the global model. Specifically, if a shortcut is injected into a particular module, it will likely cause this module to behave differently from the rest of the model. Taking Faster R-CNN as an example, let us consider its two critical components: the Region Proposal Network (RPN) and the Region-based CNN (RCNN). The RPN is responsible for generating object proposals, while the RCNN focuses on classifying these proposals and refining their locations. If an attacker implants a backdoor into the RCNN classification head, it may lead to inconsistent detection results between RPN and RCNN, such as proposals that are correctly identified by RPN but misclassified by RCNN. This inconsistency provides us with a strong signal to detect the presence of backdoors.

Building upon this insight, we further investigate whether the inconsistent module is the only one affected by the backdoor, as it determines the focus and scope of our backdoor removal efforts. To answer this question, we propose a simple yet effective verification method: we first reinitialize the inconsistent module, and then fine-tune the entire model using a small set of clean data. Intriguingly, we find that the resulting model exhibits completely consistent behavior on both clean and poisoned datasets, indicating that the inconsistent module is indeed the "Achilles' heel" of the backdoored model, where the attacker's payload is concentrated. 

\begin{algorithm}[t]
\caption{Backdoor Detection Algorithm Based on RPN and R-CNN Inconsistency}
\label{alg:backdoor_detection}
\begin{algorithmic}[1]
\Require Attacked model $f(\cdot)$, trigger sample set $\mathcal{D}_{\text{trigger}}$, negligible difference threshold $\epsilon$
\Ensure Backdoor attack judgment result
\State $\mathcal{S} \leftarrow \emptyset$ \Comment{Initialize inconsistency score set}
\For{$\mathbf{x} \in \mathcal{D}_{\text{trigger}}$}
    \State // Extract RPN output $\{\mathbf{r}_i\}_{i=1}^N$ and R-CNN output $\{(\mathbf{p}_i, \mathbf{t}_i)\}_{i=1}^N$ for $\mathbf{x}$
    \For{$i=1$ \textbf{to} $N$}
        \State // Compute classification score difference of the $i$-th proposal between RPN and R-CNN
        \State $s_i \leftarrow |\mathbf{r}_i - \mathbf{p}_i|$
        \If{$s_i > \epsilon$}
            \State // Only keep scores with significant differences
            \State $\mathcal{S} \leftarrow \mathcal{S} \cup \{s_i\}$
        \EndIf
    \EndFor
\EndFor
\State // Compute the arithmetic mean of remaining scores
\State $\mu \leftarrow \frac{1}{|\mathcal{S}|} \sum_{s \in \mathcal{S}} s$
\If{$\mu > \theta$}
    \State // Model $f(\cdot)$ is possibly attacked by backdoor
    \State \Return Model $f(\cdot)$ is possibly attacked by backdoor
\Else
    \State // Model $f(\cdot)$ is normal
    \State \Return Model $f(\cdot)$ is normal
\EndIf
\end{algorithmic}
\end{algorithm}

This observation leads to a powerful and efficient backdoor removal strategy. Instead of the complex and costly techniques used in previous works, such as pruning or fine-tuning the entire model, we can achieve effective backdoor removal by simply reinitializing the infected module and fine-tuning the model with a small clean dataset. This localized reinitialization erases the backdoor-related information in the infected module, while the global fine-tuning step allows the model to adapt to this change and maintain its performance on clean data. Through extensive experiments, we demonstrate that our method can successfully remove backdoors from a variety of object detection models, without sacrificing their accuracy on clean inputs.

\begin{algorithm}[t]
\caption{Backdoor Removal Algorithm via Local Initialization and Fine-tuning}
\label{alg:backdoor_removal}
\begin{algorithmic}[1]
\Require
\State $\mathcal{D}_{\text{clean}}$: Clean training dataset
\State $M$: Target detection model infected by backdoor
\State $E$: Number of fine-tuning epochs
\State $A$: Data augmentation method
\Ensure
\State $M_{\text{fine-tuned}}$: Fine-tuned target detection model
\State // Identify the key module affected by backdoor
\State $\text{affected\_module} \leftarrow \text{IdentifyAffectedModule}(M)$
\State // Locally initialize the affected parameters
\State $M_{\text{init}} \leftarrow \text{LocallyInitialize}(M, \text{affected\_module})$
\For{$e = 1$ to $E$}
    \For{each batch $(X, Y)$ in $\mathcal{D}_{\text{clean}}$}
        \State // Apply data augmentation
        \State $(X_{\text{aug}}, Y_{\text{aug}}) \leftarrow \text{DataAugment}(X, Y, A)$
        \State // Fine-tune the model on augmented clean data
        \State Update $M_{\text{init}}$'s parameters to minimize the loss on $(X_{\text{aug}}, Y_{\text{aug}})$
    \EndFor
\EndFor
\State $M_{\text{fine-tuned}} \leftarrow M_{\text{init}}$
\State \Return $M_{\text{fine-tuned}}$
\end{algorithmic}
\end{algorithm}

\begin{figure*}[t]
    \centering
    \begin{minipage}{0.37\textwidth}
        \centering
        \includegraphics[width=\textwidth]{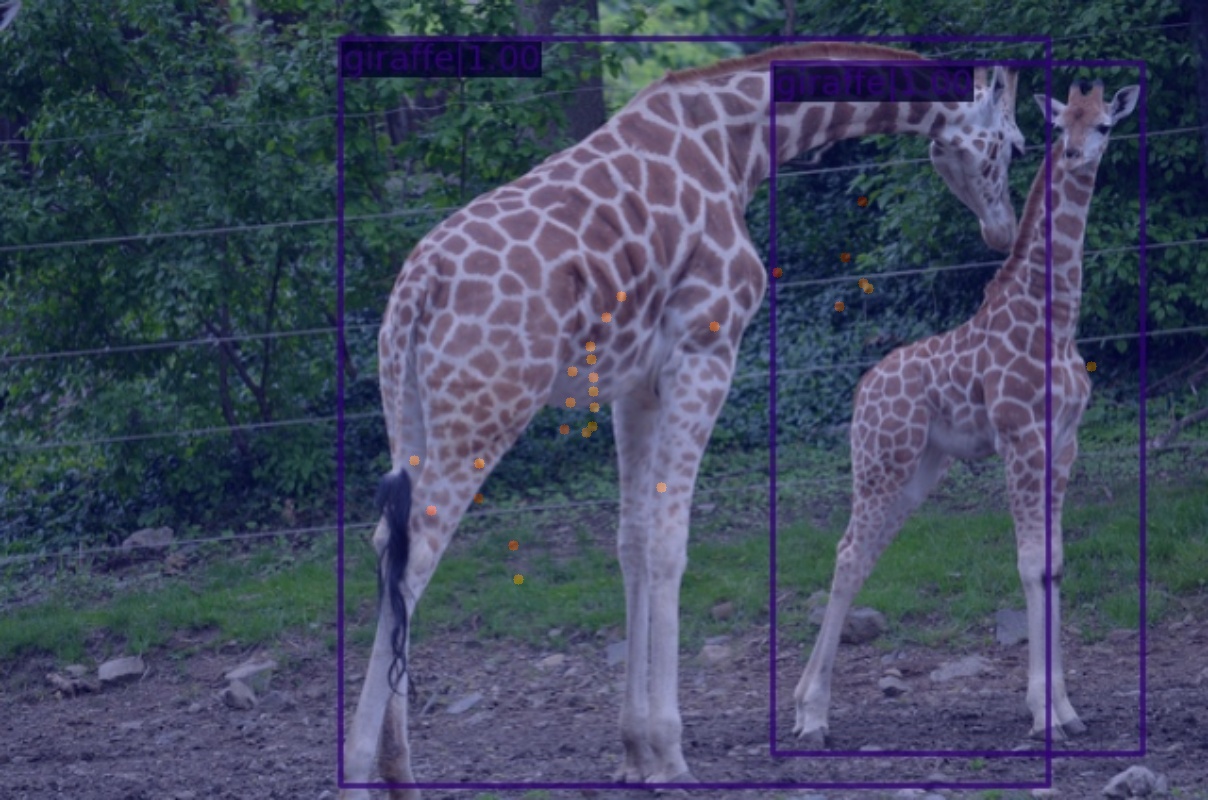}
        {\footnotesize{(a) Clean Model Inconsistency Heatmap}}
    \end{minipage}
    \hfill
    \begin{minipage}{0.37\textwidth}
        \centering
        \includegraphics[width=\textwidth]{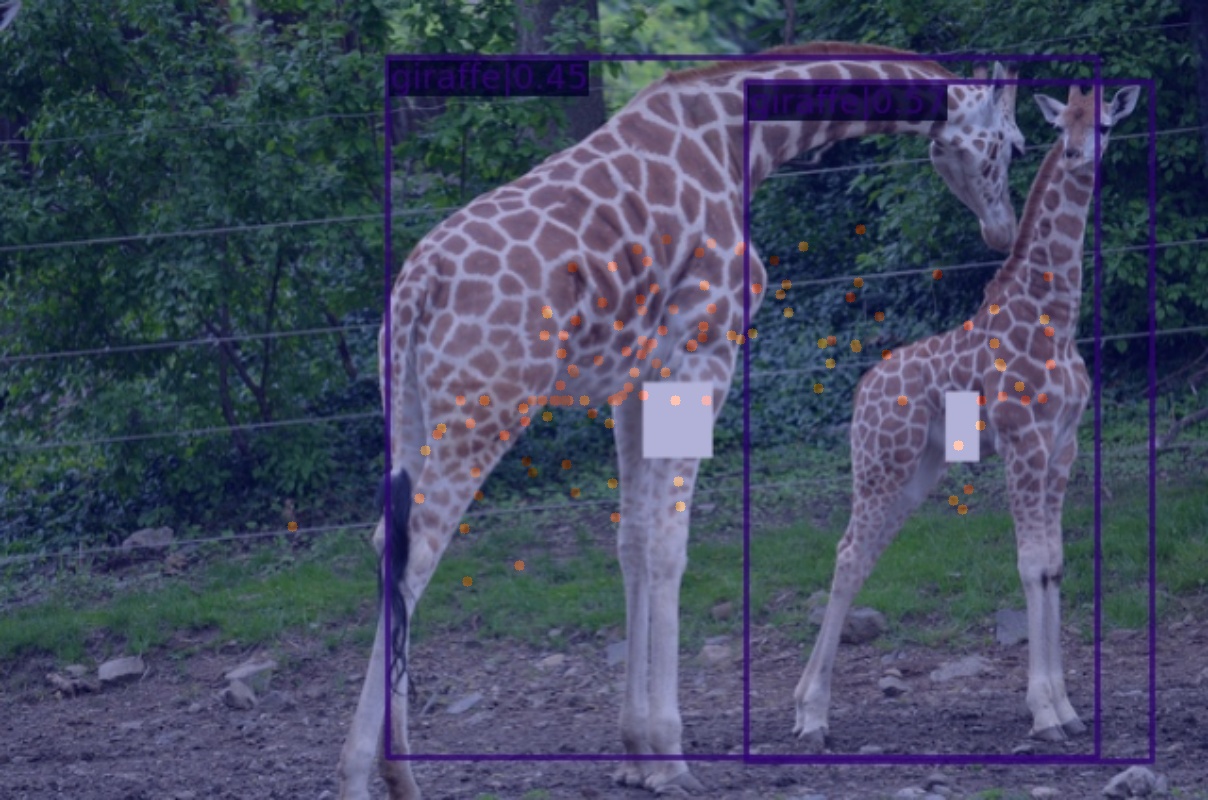}
        {\footnotesize{(b) Poisoned Model Inconsistency Heatmap}}
    \end{minipage}
    \hfill
    \begin{minipage}{0.25\textwidth}
        \centering
        \includegraphics[width=\textwidth]{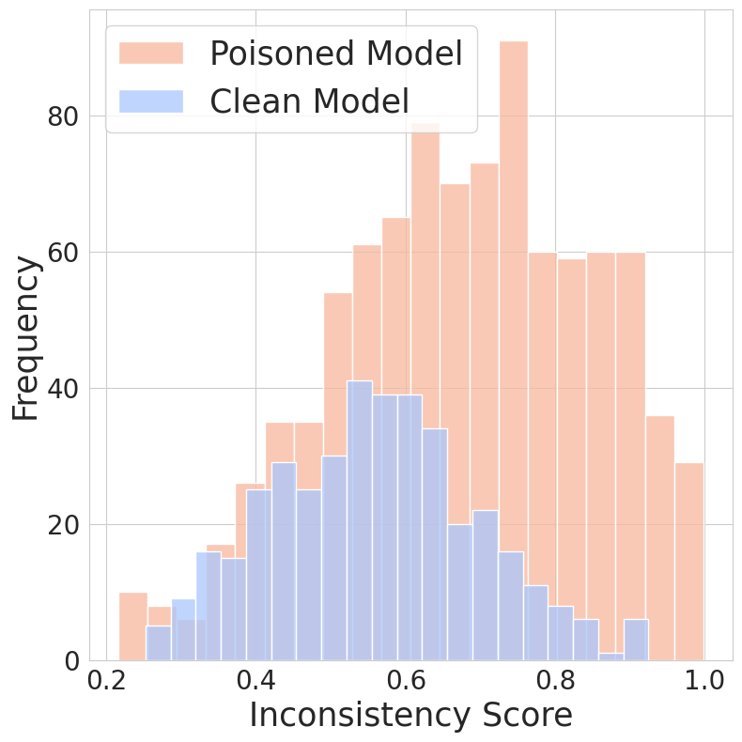}
        \footnotesize{(c) Inconsistency Heatmap Histogram}
    \end{minipage}
    \caption{ The inconsistency scores around the trigger are significantly higher than those at the corresponding locations in clean samples. As reflected in the histograms, the mean inconsistency scores of the toxic samples are greater than those of the clean samples.}
    \label{fig:inconsistency_comparison}
\end{figure*}

\subsection{Cross-Module Inconsistency Detection}

Based on the inconsistency between RPN and R-CNN, the backdoor detection algorithm consists of the following key steps:

\textbf{1. Inconsistency Score Calculation:} For each trigger sample $\mathbf{x}$ in the trigger sample set $\mathcal{D}_{\text{trigger}}$, we first extract its RPN output ${\left\{\mathbf{r}_i\right\}}_{i=1}^N$ and R-CNN output ${(\mathbf{p}_i, \mathbf{t}_i)}_{i=1}^N$, where $N$ is the number of proposals. Each $\mathbf{r}_i$ represents the RPN's classification score for the $i$-th proposal, while $\mathbf{p}_i$ and $\mathbf{t}_i$ denote the R-CNN's classification score and bounding box for the same proposal, respectively. Then, for each proposal, we compute the difference between its RPN classification score and R-CNN classification score as the inconsistency score $s_i$: $$s_i = \Vert\mathbf{r}_i - \mathbf{p}_i\Vert_1$$ Intuitively, if the model is not backdoored, the RPN and R-CNN should give consistent predictions for the same proposal, leading to a small $s_i$. However, if the model is injected with a backdoor, the trigger may cause the RPN and R-CNN to behave inconsistently, resulting in a large $s_i$.

\textbf{2. Negligible Difference Threshold Setting:}In practice, the inconsistency scores $s_i$ may be affected by various factors other than backdoors, such as the inherent discrepancy between the RPN and R-CNN, the quality of proposals, etc. To filter out these negligible differences, we introduce a negligible difference threshold $\epsilon$. Only those scores $s_i$ that are greater than $\epsilon$ are considered as significant inconsistencies and are collected into the set $\mathcal{S}$ for further analysis. The choice of $\epsilon$ depends on the specific data distribution and can be adjusted based on validation data.

\textbf{3. Arithmetic Mean Calculation:}After obtaining the set of significant inconsistency scores $\mathcal{S}$, we compute their arithmetic mean $\mu$ as: $$\mu = \frac{1}{|\mathcal{S}|}\sum_{s \in \mathcal{S}} s$$ The arithmetic mean $\mu$ serves as an overall measure of the level of inconsistency between RPN and R-CNN. A high $\mu$ indicates that the model's behavior is highly inconsistent, which is a strong signal of the presence of backdoors.

\textbf{4. Backdoor Judgment Threshold Selection:}Finally, we compare the arithmetic mean $\mu$ with a predefined backdoor judgment threshold $\theta$. If $\mu$ is greater than $\theta$, we consider the model to be possibly attacked by a backdoor; otherwise, we consider the model to be normal. The selection of $\theta$ is based on the desired trade-off between detection accuracy and false alarm rate, and can be tuned using validation data.

\begin{table}[t]
\centering
\caption{Backdoor Detection Results}
\label{tab:backdoor_detection_results}
\setlength{\tabcolsep}{4pt} 
\small 
\begin{tabular}{lccc}
\hline
\textbf{Model} & \textbf{$\mu$ (Clean)} & \textbf{$\mu$ (Poisoned)} & \textbf{Detection Result} \\ \hline
Faster R-CNN & 0.55 & 0.67 & Backdoor Detected \\
Faster R-CNN FPN & 0.49 & 0.63 & Backdoor Detected \\
Mask R-CNN & 0.51 & 0.72 & Backdoor Detected \\
Double-Head R-CNN & 0.46 & 0.61 & Backdoor Detected \\ \hline
\end{tabular}
\end{table}

\subsection{Targeted Renewal Fine-tuning}

Exploiting the insight that the inconsistent module is the primary target of backdoor injection, our backdoor removal algorithm consists of three main steps: identifying the affected module, locally initializing the affected parameters, and fine-tuning the model on augmented clean data.

\textbf{1. Identifying the Affected Module:} We first identify the key module most affected by the backdoor using the function $\text{IdentifyAffectedModule}(M)$. This function leverages the inconsistency scores computed in the backdoor detection algorithm to determine the module with the highest average inconsistency, which is considered the most likely target of the backdoor injection.

\textbf{2. Locally Initializing the Affected Parameters:} Once the affected module is identified, we perform a local initialization of its parameters using the function $\text{LocallyInitialize}(M, \text{affected\_module})$. This function resets the parameters of the affected module to random values, while keeping the parameters of other modules unchanged. This step aims to erase the backdoor influence concentrated in the affected module.

\textbf{3. Fine-tuning on Augmented Clean Data:} Finally, we fine-tune the locally initialized model $M_{\text{init}}$ on the clean training dataset $\mathcal{D}{\text{clean}}$ for $E$ epochs. In each training batch, we first apply data augmentation~\cite{mumuni2022data} to the clean data $(X, Y)$ using the augmentation method $A$, obtaining the augmented data $(X{\text{aug}}, Y_{\text{aug}})$. Then, we update the model parameters to minimize the loss on the augmented clean data. This fine-tuning process helps the model adapt to the initialized parameters and further reduces any residual backdoor effect, while maintaining its performance on normal data.

The algorithm returns the fine-tuned model $M_{\text{fine-tuned}}$ as the target detection model with the backdoor removed.

\section{Experiments}
\begin{table*}[t]
\caption{Experimental Results on the Poisoned and Clean Dataset}
\begin{tabular}{c|c|cccccc|cccccc}
\hline
\multirow{2}{*}{Model} & \multirow{2}{*}{Metric} & \multicolumn{6}{c|}{Poisoned Dataset} & \multicolumn{6}{c}{Clean Dataset} \\ \cline{3-14} 
                       &                         & AP    & AP$_{50}$ & AP$_{75}$ & AP$_{S}$ & AP$_{M}$ & AP$_{L}$ & AP    & AP$_{50}$ & AP$_{75}$ & AP$_{S}$ & AP$_{M}$ & AP$_{L}$ \\ \hline
\multirow{3}{*}{Faster R-CNN} & Original & 0.088 & 0.184 & 0.09  & 0.071 & 0.072 & 0.116 & 0.337 & 0.549 & 0.383 & 0.188 & 0.372 & 0.474 \\ \cline{2-14} 
                              & Vanilla  & 0.149 & 0.268 & 0.147 & 0.065 & 0.126 & 0.200 & 0.281 & 0.517 & 0.286 & 0.153 & 0.319 & 0.354 \\
                            & Ours & \textbf{0.285} & \textbf{0.497} & \textbf{0.263} & \textbf{0.15} & \textbf{0.29} & \textbf{0.368} & 
                            0.285 & 0.497 & 0.263 & 0.15  & 0.29  & 0.368 \\ \hline

\multirow{3}{*}{Faster R-CNN FPN} & Original & 0.095 & 0.185 & 0.087 & 0.072 & 0.074 & 0.131 & 0.367 & 0.567 & 0.393 & 0.207 & 0.404 & 0.489 \\ \cline{2-14} 
                                  & Vanilla  & 0.149 & 0.269 & 0.148 & 0.071 & 0.142 & 0.225 & 0.308 & 0.529 & 0.311 & 0.157 & 0.351 & 0.388 \\
                                  & Ours & \textbf{0.291} & \textbf{0.506} & \textbf{0.317} & \textbf{0.152} & \textbf{0.326} & \textbf{0.372} & 0.291 & 0.506 & 0.317 & 0.152 & 0.326 & 0.372 \\ \hline
\multirow{3}{*}{Mask R-CNN} & Original & 0.098 & 0.191 & 0.09  & 0.074 & 0.077 & 0.144 & 0.373 & 0.625 & 0.401 & 0.228 & 0.411 & 0.539 \\ \cline{2-14} 
                            & Vanilla  & 0.164 & 0.277 & 0.153 & 0.073 & 0.156 & 0.232 & 0.313 & 0.582 & 0.316 & 0.173 & 0.357 & 0.427 \\
                            & Ours & \textbf{0.301} & \textbf{0.521} & \textbf{0.306} & \textbf{0.166} & \textbf{0.359} & \textbf{0.382}
                             & 0.301 & 0.521 & 0.306 & 0.166 & 0.359 & 0.382 \\ \hline
\multirow{3}{*}{Double-Head R-CNN} & Original & 0.105 & 0.186 & 0.091 & 0.071 & 0.081 & 0.146 & 0.384 & 0.63  & 0.411 & 0.231 & 0.423 & 0.544 \\ \cline{2-14} 
                                   & Vanilla  & 0.171 & 0.291 & 0.152 & 0.071 & 0.164 & 0.238 & 0.323 & 0.588 & 0.326 & 0.175 & 0.368 & 0.432 \\
& Ours & \textbf{0.318} & \textbf{0.527} & \textbf{0.297} & \textbf{0.168} & \textbf{0.381} & \textbf{0.386}
 & 0.318 & 0.527 & 0.297 & 0.168 & 0.381 & 0.386 \\ \hline
\end{tabular}
\label{tab:removal_results}
\end{table*}

\begin{figure*}[t]
\centering
\begin{subfigure}{.24\textwidth}
  \centering
  \includegraphics[width=\linewidth]{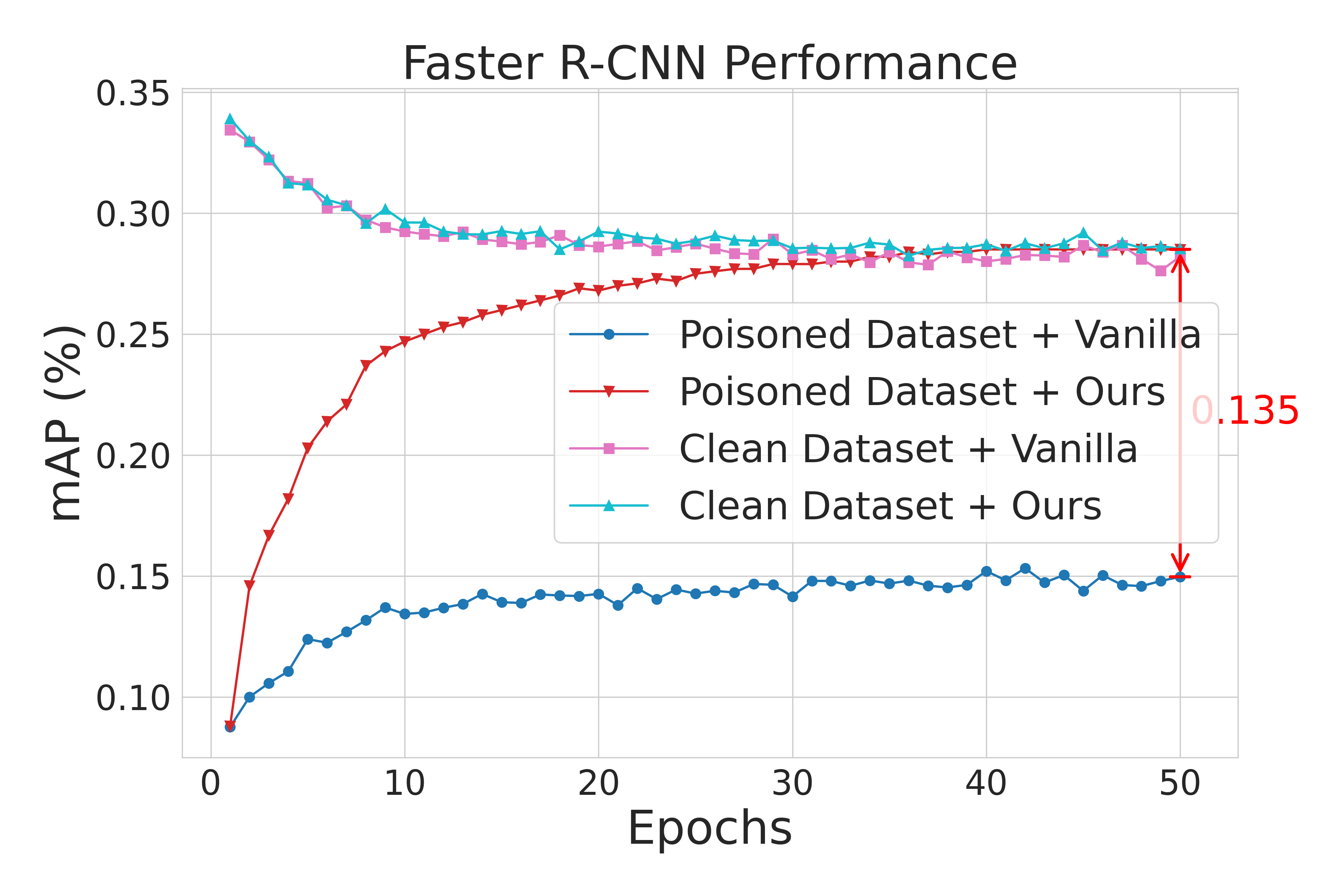}
  \caption{Faster R-CNN Without RPN}
  \label{fig:sub-first}
\end{subfigure}%
\hspace*{2mm} 
\begin{subfigure}{.24\textwidth}
  \centering
  \includegraphics[width=\linewidth]{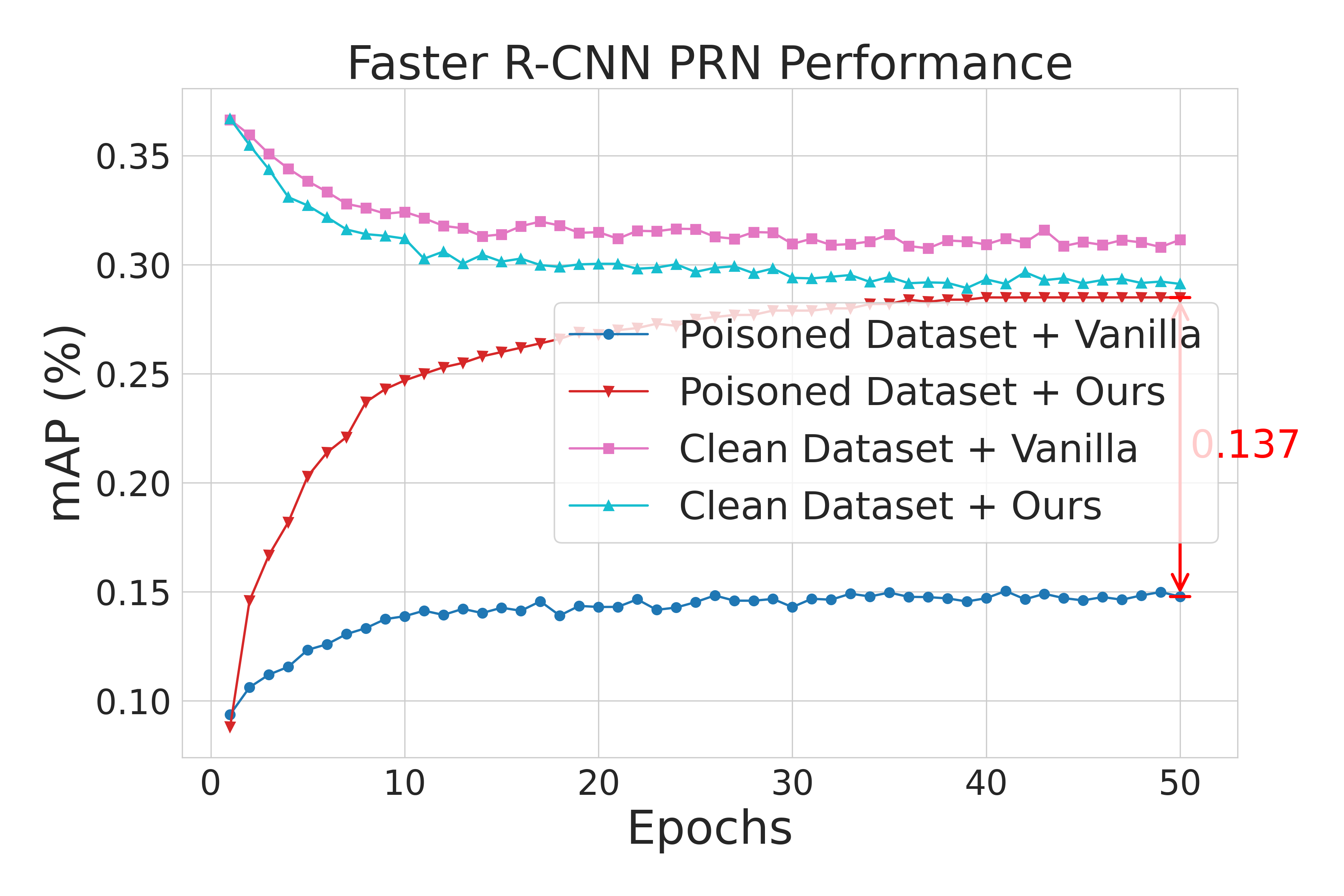}
  \caption{Faster R-CNN}
  \label{fig:sub-second}
\end{subfigure}%
\hspace*{2mm} 
\begin{subfigure}{.24\textwidth}
  \centering
  \includegraphics[width=\linewidth]{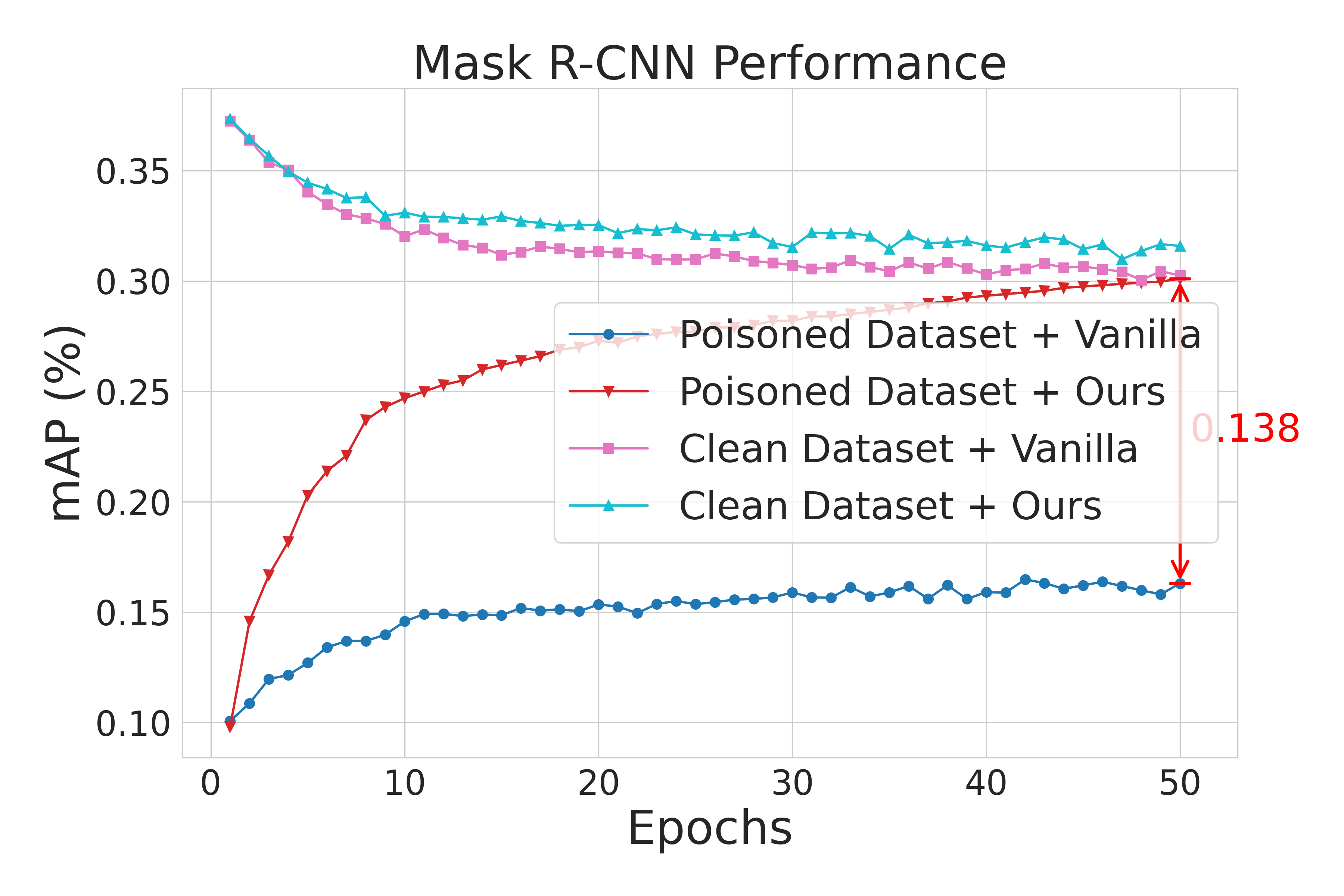}
  \caption{Mask R-CNN}
  \label{fig:sub-third}
\end{subfigure}%
\hspace*{2mm} 
\begin{subfigure}{.24\textwidth}
  \centering
  \includegraphics[width=\linewidth]{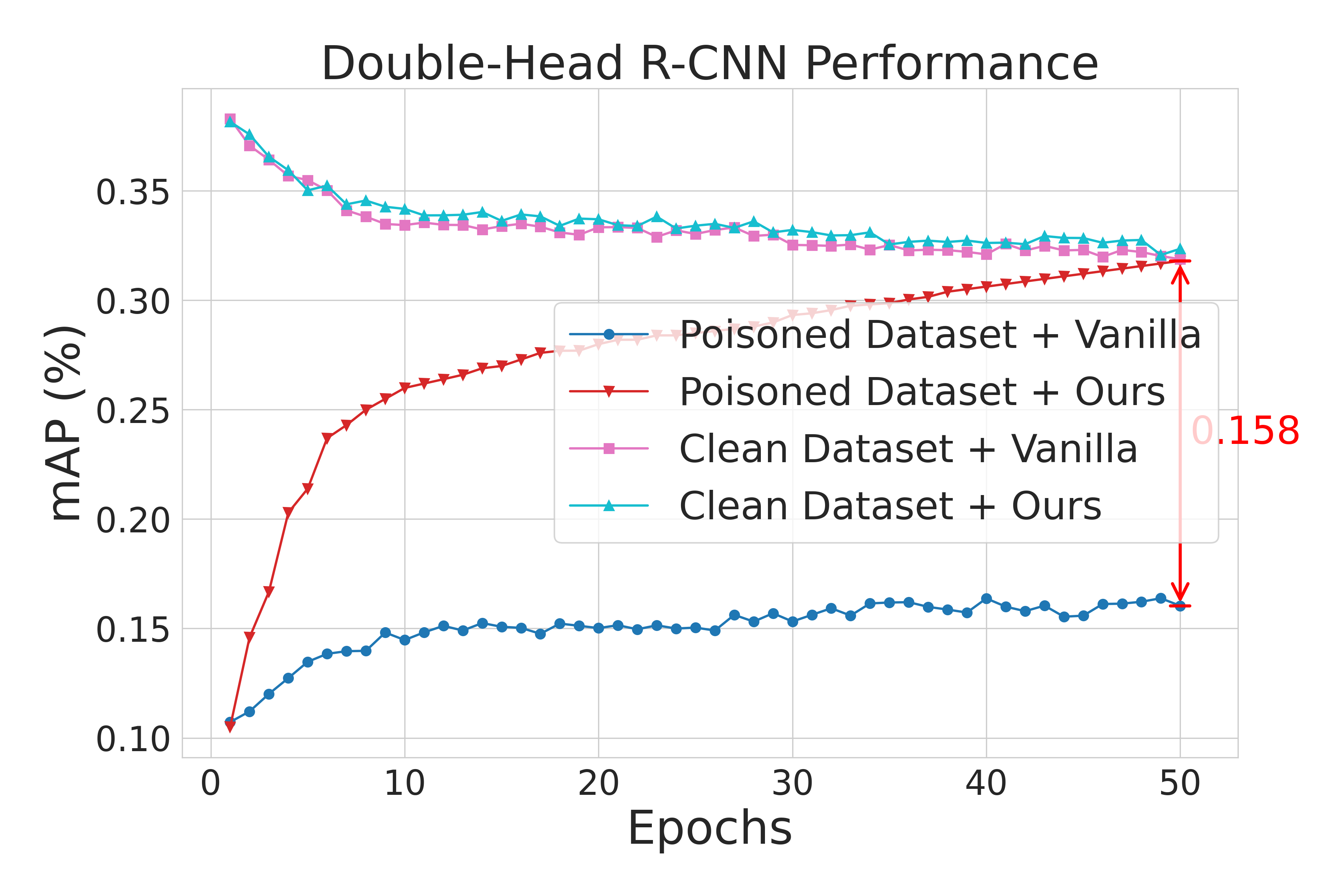}
  \caption{Double-Head R-CNN}
  \label{fig:sub-fourth}
\end{subfigure}
\caption{Performance Comparison of Backdoor Removal Methods and Naive Fine-Tuning on Clean and Poisoned Datasets}
\label{fig:rcnn-comparison}
\end{figure*}
\subsection{Experimental Settings}

\textbf{Model Structure and Dataset Description.} We adopt four representative object detectors, including Faster R-CNN, Faster R-CNN FPN, Mask R-CNN, and Double-Head R-CNN, for the evaluations. Besides, following the classical setting in object detection, we use the COCO dataset~\cite{cocodataset} as the benchmark for our discussions.

\textbf{Attack Setup.} Following the setup in~\cite{lin2014microsoft}, we simplify the approach by utilizing a white patch as the trigger pattern, with a poisoning rate established at 5\%. Consistent with the methodology outlined in~\cite{luo2023untargeted}, the dimension of the trigger for each object is configured to be 1\% of its ground-truth bounding box size, which equates to 10\% of both the width and height, positioned centrally.

\textbf{Evaluation Metric.} For our assessment criteria, we utilize six traditional metrics centered on average precision, as outlined in [20]. These include: 1) mAP, 2) $\mathrm{AP}_{50}$, 3) $\mathrm{AP}_{75}$, 4) $\mathrm{AP}_s$ (small objects), 5) $\mathrm{AP}_m$ (medium objects), and 6) $\mathrm{AP}_1$ (large objects). We compute these metrics separately across both the unaltered test dataset and its fully poisoned counterpart, the latter having a poisoning rate of 100\%. 

\textbf{Baseline.} As the first work specifically targeting backdoor defense in object detection models, we choose fine-tuning as our baseline because it is the most widely used method that does not rely on any prior knowledge or assumptions about the backdoor.

\subsection{Backdoor Detection}

To intuitively understand the impact of backdoors on the internal behavior of models, we first generate heatmaps of the inconsistency between RPN and R-CNN outputs for both clean and backdoored models, as shown in Figure \ref{fig:inconsistency_comparison} (a) and (b). By comparing the heatmaps of the two types of models, we observe that the inconsistency distribution of backdoored models is significantly higher than that of clean models. Furthermore, we plot histograms of the inconsistency scores, as depicted in Figure \ref{fig:inconsistency_comparison} (c), which further reveals the notable difference between the score distributions of backdoored and clean models. These visualizations provide us with an intuitive understanding that the presence of backdoors indeed leads to inconsistencies between the internal components of the model.

This observation aligns with our key intuition: if a backdoor is injected into a specific module of the model, the behavior of that module is likely to be inconsistent with the rest of the model. Taking Faster R-CNN as an example, if an attacker implants a backdoor into the classification head of the RCNN, it may result in inconsistent detection results between the RPN and RCNN, such as proposals correctly identified by the RPN being misclassified by the RCNN. This inconsistency provides us with a strong signal for detecting the presence of backdoors.

To further quantify the impact of backdoors on model inconsistency, we conduct experiments on four object detection models: Faster R-CNN, Faster R-CNN FPN, Mask R-CNN, and Double-Head R-CNN. For each model, we train both clean and backdoored versions using five different random initialization parameters. After setting a threshold $\theta=0.58$, we compute the average inconsistency score $\mu$ for each model on both the clean dataset and the backdoor trigger dataset. The experimental results are presented in Table \ref{tab:backdoor_detection_results}. We observe that for all backdoored models, the $\mu$ values are significantly higher than those of the clean models and exceed the threshold $\theta$. These quantitative results further confirm that our algorithm can effectively capture the internal inconsistencies caused by backdoors, thereby accurately detecting the presence of backdoors in the models.

\subsection{Backdoor Defense}
To validate the effectiveness of our proposed backdoor removal method, we conduct experiments on four widely-used object detection models: Faster R-CNN, Faster R-CNN FPN, Mask R-CNN, and Double-Head R-CNN. We evaluate the performance of these models under three scenarios: (1) Original, where the model is infected with a backdoor; (2) Vanilla, where the backdoored model is fine-tuned on a clean dataset; and (3) Ours, where the proposed backdoor removal method is applied. The experiments are performed on both a poisoned dataset, which contains the backdoor trigger, and a clean dataset without the trigger. The experimental results are presented in Table \ref{tab:removal_results} and Figure \ref{fig:rcnn-comparison}.

\begin{figure}[t]
  \centering
  \begin{minipage}[t]{0.23\textwidth}
    \centering
    \includegraphics[width=\textwidth]{img/000000050638_poisoned.jpg} 
    \caption*{(a) Poisoned Model} 
    \label{fig:children_compare_image1}
  \end{minipage}
  \hfill 
  \begin{minipage}[t]{0.23\textwidth}
    \centering
    \includegraphics[width=\textwidth]{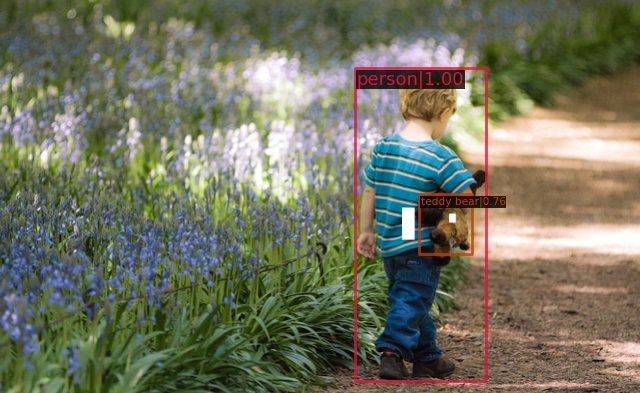} 
    \caption*{(b) Clean Model} 
    \label{fig:children_compare_image2}
  \end{minipage}
  \caption{(a) The poisoned model fails to detect the child in the presence of the trigger pattern. (b) After undergoing our inconsistency analysis and targeted reset finetuning, the model successfully detects the child even with the trigger present.}
  \label{fig: children_compare}
\end{figure}

For the Faster R-CNN model, our method achieves an AP of 0.285 on the poisoned dataset, significantly outperforming the original backdoored model (0.088) and the vanilla fine-tuning approach (0.149). This demonstrates the effectiveness of our method in removing backdoors from the model. Moreover, our method maintains an AP of 0.285 on the clean dataset, which is comparable to the performance of the vanilla fine-tuning approach (0.281) and only slightly lower than the original model's performance (0.337). This indicates that our method successfully preserves the model's performance on clean data while effectively eliminating the backdoor.

Similar trends can be observed for the other three object detection models. For Faster R-CNN FPN, our method achieves an AP of 0.291 on the poisoned dataset, surpassing both the original backdoored model (0.095) and the vanilla fine-tuning approach (0.149). On the clean dataset, our method maintains an AP of 0.291, which is close to the performance of the vanilla fine-tuning approach (0.308). For Mask R-CNN and Double-Head R-CNN, our method consistently outperforms the original backdoored models and the vanilla fine-tuning approach on the poisoned dataset, while preserving the performance on the clean dataset.

Figure \ref{fig:rcnn-comparison} presents a visual comparison of the performance trends during the backdoor removal process for each object detection model. As the number of epochs increases, our proposed method  exhibits a rapid and stable improvement in mAP scores on the poisoned dataset, significantly outperforming vanilla fine-tuning.  Moreover, our method maintains a high mAP score on the clean dataset, closely matching the performance of the model fine-tuned on the clean dataset using vanilla fine-tuning. 

The experimental results, both quantitative and visual, provide strong evidence for the effectiveness of our proposed backdoor removal method. Across all four object detection models, our method consistently outperforms vanilla fine-tuning in terms of removing backdoors and maintaining performance on clean data. The significant improvements in AP scores on the poisoned dataset, as shown in Table \ref{tab:removal_results}, demonstrate the ability of our method to neutralize the effect of the backdoor trigger. 


\begin{table}[t]
\centering
\caption{ablation study}
\begin{tabular}{
  l 
  S[table-format=1.3] 
  S[table-format=1.3] 
  S[table-format=1.3] 
  S[table-format=1.3] 
  S[table-format=1.3] 
  S[table-format=1.3]
}
\toprule
{Method/Metric}                    & {AP} & {AP$_{50}$} & {AP$_{75}$} & {AP$_{S}$} & {AP$_{M}$} & {AP$_{L}$} \\
\midrule
TRF                            & 0.262 & 0.453 & 0.268 & 0.136 & 0.298 & 0.328 \\
TRF+PD            & 0.273 & 0.472 & 0.282 & 0.145 & 0.316 & 0.350 \\
TRF+PD+RD & 0.291 & 0.506 & 0.297 & 0.152 & 0.326 & 0.372 \\
\bottomrule
\end{tabular}
\label{tab:ablation_study}
\end{table}

\subsection{Ablation Study}

To further investigate the effectiveness of different components in our proposed backdoor removal method, we conduct an ablation study on the Faster R-CNN FPN model. The results are presented in Table \ref{tab:ablation_study}. We examine three variations of our method: (1) Targeted Renewal Finetuning (TRF), (2) TRF with Photodistortion (TRF+PD), and (3) TRF with Photodistortion and Random Flip (TRF+PD+RD). Here, TRF stands for Targeted Renewal Finetuning, PD represents Photodistortion, and RD denotes Random Flip.

The baseline method, TRF, achieves an AP of 0.262, demonstrating the effectiveness of finetuning the model with targeted renewal in removing backdoors. By incorporating photodistortion during finetuning (TRF+PD), the AP improves to 0.273, indicating that the introduction of data augmentation techniques enhances the model's robustness and generalization ability. Finally, the addition of random flipping (TRF+PD+RD) further boosts the AP to 0.291, showcasing the benefits of combining multiple data augmentation strategies.



\section{Conclusion}
In this paper, we presented a novel framework for backdoor detection and removal in object detection models. We proposed to exploit the inconsistency between the behaviors of the region proposal network (RPN) and the region classification network (R-CNN) as a strong indicator of backdoor presence. We developed a simple yet effective detection algorithm based on prediction inconsistency and devised a removal strategy that localizes the backdoor removal to the affected module via parameter reinitialization and global fine-tuning. Extensive experiments on multiple state-of-the-art object detectors demonstrated the effectiveness and generality of our method, significantly outperforming baseline removal methods while maintaining high clean data accuracy. To the best of our knowledge, our work represents the first systematic study of backdoor detection and removal in object detection models, offering a principled and efficient solution to enhance the robustness and trustworthiness of these models in safety-critical applications.


\bibliographystyle{ACM-Reference-Format}
\bibliography{sample-base}

\end{document}